\documentclass[journal]{IEEEtran}
\ifCLASSINFOpdf
\else
\fi

\usepackage{graphicx}
\usepackage{multirow}
\usepackage{multicol}
\usepackage{amsmath}
\usepackage{array}
\usepackage{tabulary}
\newcolumntype{K}[1]{>{\centering\arraybackslash}p{#1}}
\usepackage{threeparttable}

\usepackage{url}
\usepackage{mathrsfs,amsfonts,amsthm,amssymb,algorithm}

\begin{document}
%
\title{Prior Knowledge Driven Label Embedding for Slot Filling in Natural Language Understanding}
%
%
%

\author{Su~Zhu,~\IEEEmembership{Student~Member,~IEEE,} Zijian~Zhao, Rao~Ma,
        and~Kai~Yu,~\IEEEmembership{Senior~Member,~IEEE}
\thanks{
 
Su Zhu, Zijian Zhao, Rao Ma and Kai Yu are supported by the National Key Research and Development Program of China (Grant No.2017YFB1002102). Experiments have been carried out on the PI supercomputer at Shanghai Jiao Tong University. (\emph{Corresponding authors}: Kai Yu.)

The authors are with the SpeechLab and MoE Key Lab of Artificial Intelligence, Department of Computer Science and Engineering, Shanghai Jiao Tong University, Shanghai 200240, China (e-mail: paul2204@sjtu.edu.cn; 1248uu@sjtu.edu.cn; rm1031@sjtu.edu.cn; kai.yu@sjtu.edu.cn).

}
}

\maketitle


\begin{abstract}
Traditional slot filling in natural language understanding (NLU) predicts a one-hot vector for each word. This form of label representation lacks semantic correlation modelling, which leads to severe data sparsity problem, especially when adapting an NLU model to a new domain. To address this issue, a novel label embedding based slot filling framework is proposed in this paper. Here, distributed label embedding is constructed for each slot using prior knowledge. Three encoding methods are investigated to incorporate different kinds of prior knowledge about slots: {\em atomic concepts, slot descriptions, and slot exemplars}. The proposed label embeddings tend to share text patterns and reuses data with different slot labels. This makes it useful for adaptive NLU with limited data. Also, since label embedding is independent of NLU model, it is compatible with almost all deep learning based slot filling models. The proposed approaches are evaluated on three datasets. Experiments on single domain and domain adaptation tasks show that label embedding achieves significant performance improvement over traditional one-hot label representation as well as advanced zero-shot approaches.
\end{abstract}

%
\IEEEpeerreviewmaketitle

\section{Introduction}
\label{sec:intro}
%
%
%
%

\IEEEPARstart{N}{owdays}, the development of mobile internet and smart devices has led to the tremendous growth of conversational dialogue systems, such as Amazon Alexa, Google Assistant, Apple Siri and Microsoft Cortana. Natural language understanding (NLU) is a key component of these systems, parsing user's utterances into the corresponding semantic representations~\cite{wang2005spoken} for certain narrow domains (e.g., \emph{booking restaurant}, \emph{searching flight}). As a main task of NLU, {\em slot filling} is typically treated as a sequence labelling problem in which contiguous sequences of words are tagged with semantic labels (slots)~\cite{wang2011semantic,he2003data,raymond2007generative,mesnil2015using,liu2016attention}.



Deep learning has achieved great success for the slot filling task in NLU~\cite{mesnil2015using,liu2016attention,yao2014spoken,vu2016sequential,kurata2016leveraging,zhu2016encoder,li2018self}, outperforming most traditional approaches~\cite{raymond2007generative,Zettlemoyer2007Online}. However, deep learning is notorious for performing poorly with limited labelled data. This is  called the {\em data sparsity} problem, which usually refers to limited data samples of feature-label pairs. In many supervised learning tasks (e.g., part-of-speech tagging, named-entity recognition), the data sparsity problem mainly lies on feature space since their label spaces are fixed. However, the NLU slot-filling task is domain-specific, and the output semantic labels are defined for all possible semantic scopes of the domain. This brings about {\em label space data sparsity}, which is an even more serious problem in NLU slot-filling task:
\begin{itemize}
    \item The label space of slot-filling task defined in each domain is distinct from others. Therefore, it is a challenge to exploit data of different domains efficiently.
    \item Narrow domains are increasingly popular in commercial dialogue systems. Growing narrow domains lead to fine-grained semantic labels (slots). For example, domain {\tt WEATHER} concerns {\tt city\_name}, while domain {\tt TRAVEL} makes a distinction between {\tt arrival\_city} and {\tt departure\_city}.
\end{itemize}


To mitigate the label space data sparsity problem, a label embedding based method for slot filling in NLU is proposed. The label embedding is driven by prior knowledge about slot definitions, which implicates relations of different semantic labels and promotes the adaptive training of slot filling. It changes the representation of each semantic label from the one-hot vector to a distributed one. For example, slots {\tt arrival\_city} and {\tt departure\_city} are orthogonal in one-hot vector space, but partially the same in the {\tt city} dimension. Label embedding implicating relations of slots could share patterns and reuse the data of associated slots. 


Unlike input word embeddings \cite{mikolov2013distributed,mikolov2013efficient} which could be trained with a large number of unlabelled text corpora, label embeddings involve prior knowledge from domain experts or developers. Three kinds of human knowledge and the corresponding encoding methods are investigated in this paper: atomic concept, slot description encoding and slot exemplar encoding. The atomic concept refers to a pulverized slot, which is smaller semantic unit than the slot, e.g., the atomic concepts of slot {\tt arrival\_city} can be {\tt destination} and {\tt city\_name}. The slot description literally interprets each slot in natural language, e.g., slot {\tt arrival\_city} can be described as ``city name of the destination''. The slot exemplar is an annotation example for each slot.  



The atomic concept is initially proposed in our previous work \cite{sz128-zhu-sigdial18}, while label is not represented as an embedding vector there. Here, we exploit label embedding independent of NLU model, which can be compatible with most deep learning-based slot-filling models. To further model the temporal dependencies of labels, a new label transition module based on label embedding is also proposed within the conditional random field (CRF)~\cite{Lafferty2001crf}. Our proposed framework is evaluated on DSTC 2\&3~\cite{henderson2013dialog} and SNIPS~\cite{2018arXiv180510190C} benchmarks and a custom dataset (AICar), outperforming all baselines significantly.

The following section discusses how our method relates to the previous works. We also introduce the slot filling task in section \ref{sec:st_slu}. Then, we describe the details of label embeddings for slot filling in section \ref{sec:why_label_emb} and \ref{sec:slot_emb}. The results of the extensive experiments are given in section \ref{sec:exp}. We conclude and give some future research directions in section \ref{sec:conclusion}.

\section{Related Work}

Standard approaches to solving the data sparsity problem of slot filling include semi-supervised learning~\cite{Tur2005Combining,celikyilmaz2016new,oyl11-lan-icassp18,zhu2018robust}, domain adaptation~\cite{daume2009frustratingly,jaech2016domain,kim2016frustratingly,Liu2017MultiDomainAL,kim2017domain,jha2018bag,goyal-etal-2018-fast,zhang2018joint,zhao2019data} and zero-shot learning~\cite{ferreira2015zero,yazdani2015model,chen2016zero,Bapna2017towards,lee2018zero,shah2019robust}. The semi-supervised learning and domain adaptation (mostly parameters sharing) methods focus on improving the coverage of feature space while not label space. The zero-shot learning methods involve ontological descriptions to enhance the slot filling. However, Ferreira et al. \cite{ferreira2015zero} and Yazdani et al. \cite{yazdani2015model}  made an assumption that all possible values of each slot are known, which is not practical for real-world applications. Our methods seek to interpret semantic labels (slots) in multiple dimensions where relations of slots can be inferred implicitly. It helps data reuse for different labels and domains.

Label embeddings for SLU have been introduced in several works. Chen et al. \cite{chen2016zero} exploit label embedding for intent detection but not suitable for slot filling. Some works choose to extract label embeddings from data samples \cite{kim-etal-2015-new,ma2016label,lee2018coupled,wu2019joint} by exploiting words of values tagged with a semantic label, but they only focus on the corresponding values. We consider both the value and context information in the slot exemplar encoding. Concept tagger~\cite{Bapna2017towards} and zero-shot adaptive transfer~\cite{lee2018zero} also incorporate slot embeddings from slot descriptions for the slot filling. However, they proposed a slot-independent conditional layer to predict the existence for each slot one by one, which would take more time for training and inference. Moreover, it is not possible for modelling the time series dependence of different slots in those methods. We propose label embeddings for slot filling without changing the basic workflow, which overcomes the above issues.

\section{Slot Filling in NLU}
\label{sec:st_slu}

\begin{figure}[htbp]
\centering
\includegraphics[width=0.48\textwidth]{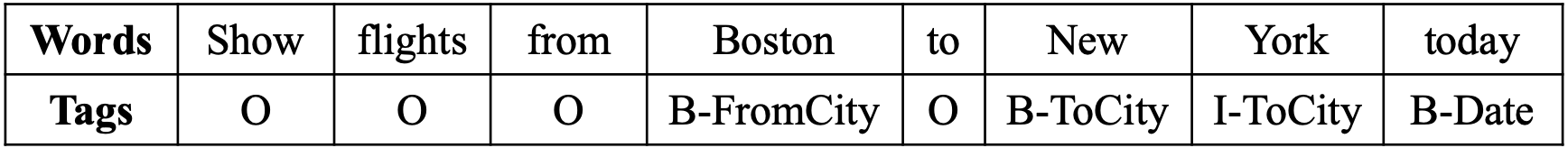}
\caption{An example of annotation for slot filling in domain \emph{searching flight}.}
\label{fig:atis}
\end{figure}

Slot filling is a major task of natural language understanding (NLU) in task-oriented dialogue systems, which automatically extracts a set of attributes or ``slots'', and the corresponding values. It is typically treated as a sequence labelling problem. An example of data annotation is provided in Fig. \ref{fig:atis}. It follows the popular inside/outside/beginning (IOB) schema, where \emph{Boston} and \emph{New York} are the departure and arrival cities specified as the slot values in the user's utterance, respectively. In this paper, we use the word \emph{tag} to refer to the semantic label in IOB schema, which is different from \emph{slot}.


In this section, we will sequentially introduce the slot filling in NLU for single domain and domain adaptation. 

\subsection{Single Domain}

Let $\boldsymbol{w} = (w_1, w_2, \cdots, w_N)$ denote an input sequence (words), and $\boldsymbol{t} = (t_1, t_2, \cdots, t_N)$ denote its output sequence (tags) in domain $d$, where $N$ is the sequence length and $t_i$ is one of the predefined tags, i.e. $t_i \in \mathcal{T}_d$. Let $\mathcal{T}_d$ denote all possible labels of domain $d$. $t_i$ would be represented as a one-hot vector, $\mathbf{o}(t_i)$, in conventional methods. The slot filling in NLU is to estimate $p(\boldsymbol{t}|\boldsymbol{w})$, the posterior probability of output sequence $\boldsymbol{t}$ given input $\boldsymbol{w}$.

Recently, motivated by a number of successful neural network and deep learning methods in natural language processing, many neural network architectures have been applied in this task, such as vanilla recurrent neural network (RNN) ~\cite{mesnil2015using,mesnil2013investigation,yao2013recurrent,vu2016bi}, convolutional neural network (CNN) ~\cite{vu2016sequential,xu2013convolutional}, long short-term memory (LSTM)~\cite{yao2014spoken,zhangA,hakanni-tur2016multidomain,reimers2017optimal}, encoder-decoder~\cite{zhu2016encoder,liu2016attention,zhai2017neural,kurata2016leveraging}. Bi-directional LSTM based RNN (BLSTM) is adopted in this paper. Every input word is mapped to a vector via $\mathbf{w}_i=\mathbf{W}_{in}\mathbf{o}(w_i)$, where $\mathbf{W}_{in}$ is an embedding matrix and $\mathbf{o}(w_i)$ a one-hot vector. Hidden vectors are recursively computed at the $i$-th time step via:
$$\mathbf{h}_i=[\overrightarrow{\mathbf{h}}_i;\overleftarrow{\mathbf{h}}_i];\overrightarrow{\mathbf{h}}_i=\text{f}_\text{{lstm}}(\mathbf{w}_i,\overrightarrow{\mathbf{h}}_{i-1});\overleftarrow{\mathbf{h}}_i=\text{b}_\text{{lstm}}(\mathbf{w}_i,\overleftarrow{\mathbf{h}}_{i+1})$$
where $[\cdot;\cdot]$ denotes vector concatenation, $\mathbf{h}_{i} \in \mathbb{R}^{2n}$ ($n$ is the hidden size), $\text{f}_\text{{lstm}}$ and $\text{b}_\text{{lstm}}$ are the LSTM functions for forward and backward passes respectively. At each time step, a linear output layer is applied to predict tags, i.e. 
\begin{equation}
\mathbf{z}_i=\mathbf{W}_o \mathbf{h}_i
\label{eqn:output_layer}
\end{equation}
where $\mathbf{W}_o \in \mathbb{R}^{|\mathcal{T}_d|\times 2n}$ denotes trainable parameters of the output layer (bias is omitted). In the following sections, prior knowledge driven label embeddings will be included in $\mathbf{W}_o$.

\subsubsection{BLSTM-softmax}
The posterior probability is factorized: 
\begin{equation}
p(\boldsymbol{t}|\boldsymbol{w}) = \prod_{i=1}^N p(t_i|\mathbf{h}_i) = \prod_{i=1}^N \text{softmax}(\mathbf{z}_i)^{\top} \mathbf{o}(t_i)
\end{equation}
where $\mathbf{o}(t_i)$ is a one-hot vector with a dimension for tag $t_i$, so $\text{softmax}(\mathbf{z}_i)^{\top} \mathbf{o}(t_i)$ is precisely the $t_i$-th element of the distribution defined by the \emph{softmax} function.


\subsubsection{BLSTM-CRF}
To model the time series dependence of output labels, several methods based on encoder-decoder architectures~\cite{kurata2016leveraging,zhu2016encoder,liu2016attention,zhai2017neural} are proposed for slot filling. Besides that, CRF output layer can jointly decode the best chain of labels for a given input sentence~\cite{yao2014recurrent,huang2015bidirectional,ma-hovy-2016-end}. The CRF output layer calculates the posterior probability at the sentence level:
\begin{equation}
p(\boldsymbol{t}|\boldsymbol{w}) = \frac{\text{exp}(\psi(\boldsymbol{w},\boldsymbol{t}))}{\sum_{\boldsymbol{t}'}\text{exp}(\psi(\boldsymbol{w},\boldsymbol{t}'))}
\end{equation}
The score of a sentence $\boldsymbol{w}$ along with a path of tags $\boldsymbol{t}$ is given by the sum of transition scores and network scores:
\begin{equation}
\psi(\boldsymbol{w},\boldsymbol{t}) = \sum_{i=1}^N ([\mathbf{A}]_{t_{i-1},t_i} + \mathbf{z}_i^{\top} \mathbf{o}(t_i))
\label{eqn:crf_score}
\end{equation}
where $\mathbf{A} \in \mathbb{R}^{|\mathcal{T}_d|\times |\mathcal{T}_d|}$ is a transition matrix, and its element $[\mathbf{A}]_{m,n}$ modelling the transition from the $m$-th to the $n$-th label for a pair of consecutive time steps.

The maximum likelihood estimation is used for training, and the negative log-likelihood is given by 
\begin{equation*}
Loss = -\log p(\boldsymbol{t} | \boldsymbol{w})
\end{equation*}
. Decoding is to search for the label sequence $\boldsymbol{t}^{*}$ with the highest posterior probability, i.e. 
\begin{equation*}
\boldsymbol{t}^{*}=\underset{\boldsymbol{t}'} {\operatorname{argmax}} \ p(\boldsymbol{t}' | \boldsymbol{w})
\end{equation*}
The searching is time independent for the \emph{softmax} output function, while it can be solved efficiently by adopting the Viterbi algorithm for the CRF output layer.

\subsection{Domain Adaptation}
\label{subsec:domain_adaptation}

Domain adaptation can alleviate the data sparsity problem of slot filling by taking advantage of data from other domains, such as feature augmentation~\cite{daume2009frustratingly}, parameters sharing~\cite{jaech2016domain,kim2016frustratingly,Liu2017MultiDomainAL}, and bag of experts~\cite{kim2017domain,jha2018bag}.

Sharing parameters for different domains is a popular approach to transferring knowledge learned from source domains to a target domain. Suppose we have a set of source domains, $\mathcal{D}_{src}$, and a target domain $td$. We mix all data of $\mathcal{D}_{src}$ to pre-train a slot filling model, then fine-tune the model by using data of $td$. The pre-training procedure can be implemented only once for different target domains.


In the pre-training stage, different domains in $\mathcal{D}_{src}$ may not be able to share the output layer fully. Because each narrow domain has a closed and limited semantic space which is different from others. It may lead to conflicts of data annotation. Several examples of disputes between different domains (e.g., DSTC2 V.S. DSTC3) are shown in Table \ref{tab:data_conflict}.

\begin{table} 
\begin{center}
\caption{Examples about conflicts of data annotation from different domains. We use \emph{[value:slot]} for annotation.}
\begin{tabular}{|l|l|}
\hline
\bf domain & \bf data samples \\
\hline
DSTC2 & I'm looking for a [Thai:Food] restaurant. \\ 
DSTC3  & I'm looking for a [Thai:Food] [restaurant:Type]. \\
\hline  
Weather & I'm going to [Beijing:City], and what's the weather?  \\ 
Travel  & I'm going to [Beijing:ToCity], and what's the weather? \\
\hline
\end{tabular}
\label{tab:data_conflict}
\end{center}
\end{table}

Therefore, the output layer should be masked at different positions, according to which domain the training data is from. For all the source domains, we have a union of labels, $\mathcal{T}_{src}=\bigcup\limits_{d\in \mathcal{D}_{src}}\mathcal{T}_d$. Therefore, the output matrix is scaled up to $\mathbf{W}_o^{src} \in \mathbb{R}^{|\mathcal{T}_{src}|\times 2n}$, and the transition matrix is $\mathbf{A}_{src} \in \mathbb{R}^{|\mathcal{T}_{src}|\times |\mathcal{T}_{src}|}$. For each domain $d \in \mathcal{D}_{src}$, there is mask vector $\mathbf{m}_d \in \mathbb{R}^{|\mathcal{T}_{src}|}$:
\begin{equation}
[\mathbf{m}_d]_{t_j}=\left\{\begin{array}{ll}{0,} & {\text { if\ } t_j \in \mathcal{T}_d} \\ {-\infty,} & {\text { otherwise }}\end{array}\right.
\end{equation}
where each $t_j \in \mathcal{T}_{src}$ is covered. It is used to control the output for domain $d$ by ignoring all labels out of this domain and replacing Equation (\ref{eqn:output_layer}) with:
\begin{equation}
\mathbf{z}_i=\mathbf{W}_o^{src} \mathbf{h}_i + \mathbf{m}_d
\label{eqn:output_layer_of_domain_adaptation}
\end{equation}

At the beginning of the fine-tuning stage of the target domain $td$, parameters are copied from the model pre-trained in the source domains for initialization. The word embedding matrix and BLSTM parameters are retained, while the output and transition matrices are copied partially:
\begin{align}
[\mathbf{W}_{o}^{td}]_{t_j}&=\left\{\begin{array}{ll}{[\mathbf{W}_{o}^{src}]_{t_j},} & {\text {if\ } t_j \in \mathcal{T}_{src} \cap \mathcal{T}_{td} } \\ {\text{random init.},} & {\text {otherwise}}\end{array}\right. \label{eqn:output_layer_sharing}\\
[\mathbf{A}_{td}]_{t_j,t_k}&=\left\{\begin{array}{ll}{[\mathbf{A}_{src}]_{t_j,t_k},} & {\text {if\ } \{t_j,t_k\} \subset \mathcal{T}_{src} \cap \mathcal{T}_{td}} \\ {\text{random init.},} & {\text {otherwise}}\end{array}\right.
\end{align}
where $t_j \in \mathcal{T}_{td}$, $t_k \in \mathcal{T}_{td}$, $\mathbf{W}_o^{td} \in \mathbb{R}^{|\mathcal{T}_{td}|\times 2n}$, $\mathbf{A}_{td} \in \mathbb{R}^{|\mathcal{T}_{td}|\times |\mathcal{T}_{td}|}$, $[\cdot]_{j}$ denotes the $j$-th row of a matrix or vector.


\section{Why Label Embedding?}
\label{sec:why_label_emb}

As illustrated in Equation (\ref{eqn:output_layer}), there is a score estimates how possible each label $t_j$ is to appear at the $i$-th time step,
\begin{equation}
[\mathbf{z}_i]_{t_j}=(\mathbf{W}_o \mathbf{h}_i)^{\top}\mathbf{o}(t_j)
\label{eqn:output_score}
\end{equation}
, before the \emph{softmax} or CRF function. 

If we consider an alternate low-rank decomposition of $\mathbf{W}_{o}$, i.e. $\mathbf{W}_{o}=\mathbf{B}^{\top} \mathbf{C}$ where $\mathbf{W}_o \in \mathbb{R}^{|\mathcal{T}_d|\times 2n}$, $\mathbf{B} \in \mathbb{R}^{K \times |\mathcal{T}_d|}$, and $\mathbf{C} \in \mathbb{R}^{K \times 2n}$. Equation (\ref{eqn:output_score}) can be rewritten as
\begin{equation}
[\mathbf{z}_i]_{t_j}=(\mathbf{B}^{\top} \mathbf{C} \mathbf{h}_i)^{\top}\mathbf{o}(t_j)=(\mathbf{C} \mathbf{h}_i)^{\top}(\mathbf{B} \mathbf{o}(t_j))
\end{equation}
$\mathbf{B}$ is a matrix corresponding to the lookup table of label embeddings, and $\mathbf{C}$ is a matrix of linear transformation, which respectively map the label $t_i$ and features $\mathbf{h}_i$ into a joint space.


Imagining that there are already pre-trained or pre-defined label embeddings which reflect relations (similarities and differences) among labels, the model learned from labelled data may even be adapted to labels never exist in the data. It can help reduce the size of required in-domain data, enabling developers to bootstrap new domains quickly. 



\section{Label Embedding for Slot Filling}
\label{sec:slot_emb}

In this section, we will introduce the proposed slot-filling method based on label embeddings, representing each output tag as a distributed vector. However, unlike word embeddings \cite{mikolov2013distributed,mikolov2013efficient} which can be pre-trained with a large number of unlabelled text corpora, data with semantic labels is very limited and in the face of the label space sparsity issue. We derive label embeddings from semantic definitions of certain domains. Three kinds of prior knowledge about slot definitions are investigated as well as the corresponding encoding methods: atomic concept, slot description and slot exemplar.


Since the label (tag) of slot filling is in IOB format, we make a simple notation to split tag into IOB and slot. For each tag $t$ (e.g., ``B-from\_city''), $\text{IOB}(t)$ gets the IOB symbol (e.g., ``B''), and $\text{SLOT}(t)$ returns the slot name (e.g., ``from\_city''). Note that $\text{SLOT}(\text{``O''})$ returns null.

\subsection{Atomic Concept}

Atomic concept assumes that there are substructures with smaller semantic units for semantic slots. It is a pure prior knowledge that each slot is represented as a set of atoms. The atomic concepts are pulverized slots and much smaller semantic units than the slots. Each slot is composed of different atomic concepts, e.g., slot {\tt from\_city} can be defined as a set of atoms \{\emph{from\_location}, \emph{city\_name}\}, and slot {\tt date\_of\_birth} can be represented as \{\emph{date}, \emph{birth}\}, as shown in Table \ref{tab:atomic_concept_example}. 

\begin{figure}[]
\centering
\includegraphics[width=0.3\textwidth]{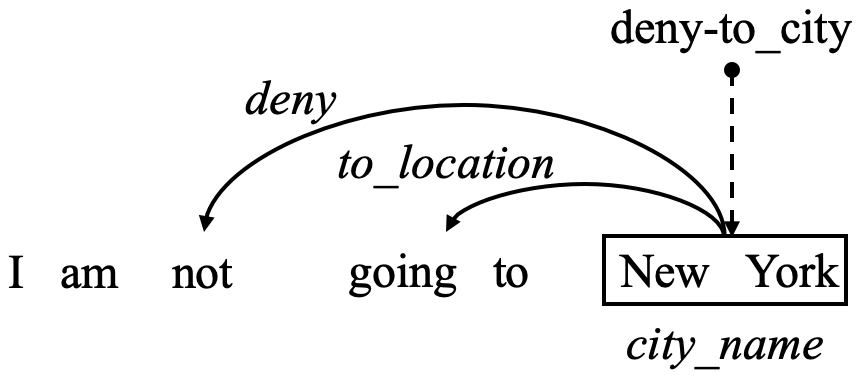}
\caption{An example of context dependency of slot {\tt deny-to\_city}. \emph{deny} and \emph{to\_location} are context-aware atomic concepts, and \emph{city\_name} focuses on the value ``New York''.}
\label{fig:atom_context_value}
\end{figure}

Atomic concepts are not automatically defined, which are designed manually by domain experts or developers. Atomic concepts can be classified into two categories, one is value-aware, and the other is context-aware. The value-aware atom is often a category name of values, e.g., ``New York'' is \emph{city\_name}. Context-aware atoms reflect implicit dependencies around the targeted value, e.g., \emph{to\_location} depends on context like ``going to'', ``fly to'', ``arrive at'', as shown in Fig. \ref{fig:atom_context_value}. Modelling on the atomic concept helps find out the linguistic patterns of related slots by atom sharing, and it can even decrease the required amount of data. 

\begin{table}
\begin{center}
\caption{\label{tab:atomic_concept_example} Examples of slot representation by atomic concepts.}
\begin{tabular}{|l|l|}
\hline \bf slot & \bf atomic concepts \\ \hline
city &  \{\emph{city\_name}\} \\
from\_city &   \{\emph{city\_name}, \emph{from\_location}\} \\
depart\_city &   \{\emph{city\_name}, \emph{from\_location}\} \\
arrive\_airport &   \{\emph{airport\_name}, \emph{to\_location}\} \\
city\_of\_birth & \{\emph{city\_name}, \emph{birth}\} \\
date\_of\_birth & \{\emph{date}, \emph{birth}\} \\
deny-to\_city & \{\emph{city\_name}, \emph{to\_location}, \emph{deny}\} \\
\hline
\end{tabular}
\end{center}
\end{table}

\subsubsection{Single Domain}

\begin{figure*}
\centering
\includegraphics[width=0.8\textwidth]{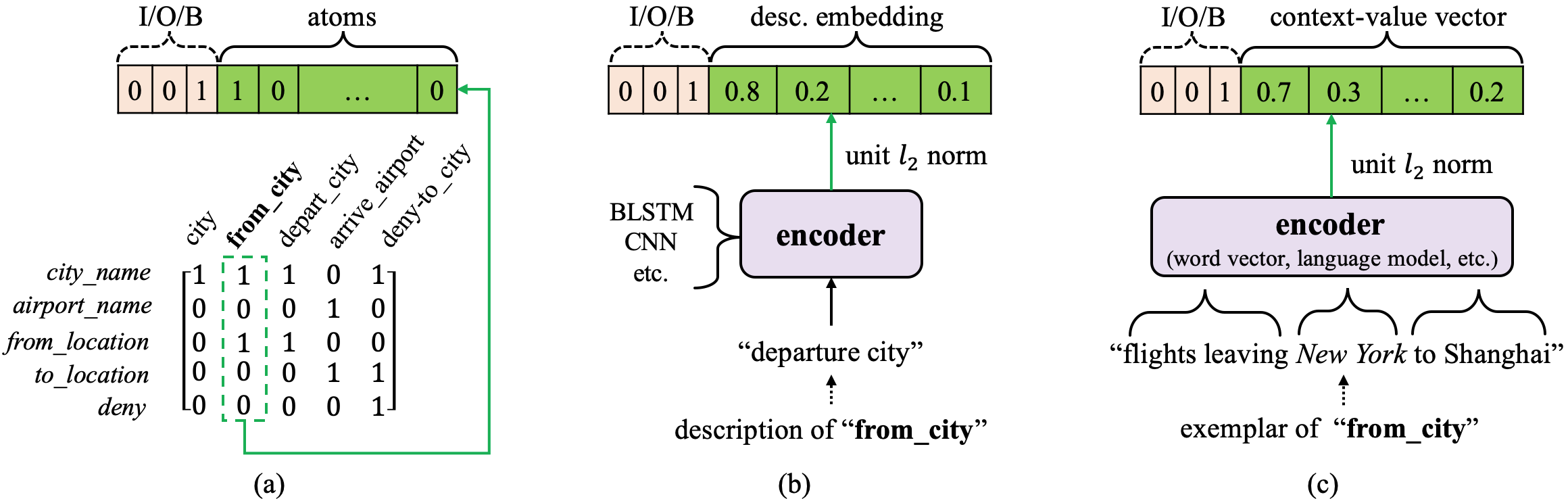}
\caption{Three different label embeddings consisted of IOB symbol and slot vector. (a) Atomic concept: each slot is represented as finer-grained semantic units. (b) Slot description encoding: each slot has a textual description in natural language which can be encoded into a vector by several sequence models. (c) Slot exemplar encoding: given a number of exemplars for each slot, the exemplars are encoded into vectors which contain the context and value information.}
\label{fig:label_embeddings}
\end{figure*}

We represent each slot $s$ of domain $d$ as a set of atomic concepts, i.e. $\text{atoms}(s)=\{a_{1},\cdots,a_{M_s}\}$, $a_{l} \in \mathcal{A}_d$, where $\mathcal{A}_d$ is the vocabulary of atomic concepts and $M_s$ is the number of atomic concepts belong to this slot. Thus, each slot can be viewed as a binary vector, $\mathbf{b}(\text{atoms}(s))$ with length of $|\mathcal{A}_d|$, while $\mathbf{b}(\text{atoms}(\text{SLOT}(\text{``O''})))=\mathbf{0}$. Due to the IOB schema, the label embedding consists of two parts, 3-dimensional IOB one-hot vector and a binary vector of atoms, as illustrated in Fig. \ref{fig:label_embeddings} (a). The label embedding of tag $t$ is
\begin{equation}
    \text{emb}(t) = [\mathbf{o}(\text{IOB}(t));\mathbf{b}(\text{atoms}(\text{SLOT}(t)))]
\end{equation}
where $\text{emb}(t) \in \mathbb{R}^{3+|\mathcal{A}_d|}$, $t \in \mathcal{T}_d$.

Finally, a label embedding matrix for domain $d$ is created, i.e., $\mathbf{B}_d \in \mathbb{R}^{(3+|\mathcal{A}_d|) \times |\mathcal{T}_d|}$, where $[\mathbf{B}_d^{\top}]_{t}=\text{emb}(t)$. Equation (\ref{eqn:output_layer}) can be rewritten as 
\begin{equation}
    \mathbf{z}_i=\mathbf{B}_d^{\top} \mathbf{C}_d \mathbf{h}_i
    \label{eqn:output_layer_atom}
\end{equation}
where $\mathbf{C}_d \in \mathbb{R}^{(3+|\mathcal{A}_d|) \times 2n}$ denotes trainable parameters of the output layer, but $\mathbf{B}_d$ is fixed as a kind of prior knowledge.


\subsubsection{Domain Adaptation} Similar to subsection \ref{subsec:domain_adaptation}, the domain adaptation with atomic concepts is also introduced as follows. For all the source domains, we have a union of atoms, $\mathcal{A}_{src}=\bigcup\limits_{d\in \mathcal{D}_{src}}\mathcal{A}_d$. Therefore, the output matrix is scaled up to $\mathbf{C}_{src} \in \mathbb{R}^{(3+|\mathcal{A}_{src}|)\times 2n}$, and $\mathbf{B}_d \in \mathbb{R}^{(3+|\mathcal{A}_{src}|) \times |\mathcal{T}_d|}$. For each domain $d \in \mathcal{D}_{src}$, there is mask vector $\mathbf{m}_d \in \mathbb{R}^{3+|\mathcal{A}_{src}|}$,
\begin{equation}
[\mathbf{m}_d]_{a_j}=\left\{\begin{array}{ll}{0,} & {\text { if\ } a_j \in \mathcal{A}_d \text{ or } a_j \in \text{IOB}} \\ {-\infty,} & {\text { otherwise }}\end{array}\right.
\end{equation}
for each $a_j \in \mathcal{A}_{src} \cup \text{IOB}$. It is used to ignore the atomic concepts out of domain $d$ and replace Equation (\ref{eqn:output_layer_of_domain_adaptation}) with:
\begin{equation}
\mathbf{z}_i=\mathbf{B}_d^{\top} (\mathbf{C}_{src} \mathbf{h}_i + \mathbf{m}_d)
\end{equation}

For fine-tuning in the target domain $td$, the word embedding matrix and BLSTM parameters are still retained. The initialization of $\mathbf{C}_{td} \in \mathbb{R}^{(3+|\mathcal{A}_{td}|)\times 2n}$ for $td$ is
\begin{align*}
[\mathbf{C}_{td}]_{a_j}&=\left\{\begin{array}{ll}{[\mathbf{C}_{src}]_{a_j},} & {\text { if\ } a_j \in (\mathcal{A}_{src}} \cap \mathcal{A}_{td}) \cup \text{IOB} \\ {\text{Random init.},} & {\text { otherwise }}\end{array}\right.
\end{align*}
for each $a_j \in \mathcal{A}_{td} \cup \text{IOB}$. Thus, we calculate output scores of the target domain as $\mathbf{z}_i=\mathbf{B}_{td}^{\top}\mathbf{C}_{td} \mathbf{h}_i$, where $\mathbf{B}_{td} \in \mathbb{R}^{(3+|\mathcal{A}_{td}|) \times |\mathcal{T}_{td}|}$.

Composition of atomic concepts for each slot provides an explicit interpretation about slot relations. However, it requires a large number of human efforts to design atomic concepts carefully, and it is hard to keep them universal for different domains.

\subsection{Slot Description Encoding}

In this method, we assume that there is a textual description for each slot in natural language. It is also pre-defined by domain experts or developers but with less human labors than the atomic concept. For example, {\tt deny-to\_city} can be described as ``not this arrival city''. The descriptions of relative slots may include same, synonymous and even antonymous words, which can help reveal the relation between different slots. Thus, we can extract label embeddings literally from slot descriptions.

Unlike the label embedding bound to atomic concepts, an encoder is required to convert literal slot descriptions of variable lengths into fixed dimension vectors, as delineated in Fig. \ref{fig:label_embeddings} (b). Suppose each slot $s$ has a description, $\text{desc}(s)=x_1 \cdots x_{L_s}$, where $L_s$ is the sequence length. Three encoders are applied: simple mean operation of word embeddings, BLSTM and CNN models. We will introduce them in a single domain, then clarify how to make domain adaptation with slot description-based label embeddings.

\subsubsection{Mean Encoder of Word Embeddings} Every description word is mapped to an $m$-dimensional vector via $\mathbf{x}_l=\mathbf{W}_{in}\mathbf{o}(x_l)$, where $\mathbf{W}_{in}$ is the word embedding matrix and $\mathbf{o}(x_l)$ a one-hot vector. The embedding of slot $s$ is simply the mean of word embeddings of all words in $\text{desc}(s)$, i.e.
\begin{equation}
    \text{mean}(\text{desc}(s)) = unit(\frac{1}{L_s} \sum_{l=1}^{L_s} \mathbf{x}_l)
\end{equation}
where $unit(\mathbf{v})=\frac{\mathbf{v}}{||\mathbf{v}||}$ gets the unit $l_2$ norm of vector $\mathbf{v}$, which enforces slot embeddings in a limited scale.

\subsubsection{BLSTM Encoder} The hidden vectors of BLSTM encoder are recursively computed at the $l$-th time step via: 
\begin{align}
\overrightarrow{\mathbf{v}}_l=&\text{f}_\text{{lstm}}(\mathbf{x}_l, \overrightarrow{\mathbf{v}}_{l-1}), l=1,\cdots,L_s\\
\overleftarrow{\mathbf{v}}_l=&\text{b}_\text{{lstm}}(\mathbf{x}_l, \overleftarrow{\mathbf{v}}_{l+1}), l=L_s,\cdots,1
\end{align}
where $\text{f}_\text{{lstm}}$ and $\text{b}_\text{{lstm}}$ are the LSTM functions for forward and backward passes respectively. We concatenate the final hidden vectors of bi-direction to be the slot embedding, i.e. 
\begin{equation}
    \text{BLSTM}(\text{desc}(s)) = unit([\overrightarrow{\mathbf{v}}_{L_s};\overleftarrow{\mathbf{v}}_1])
\end{equation}

\subsubsection{CNN Encoder} CNN is also a typical model to encode a sequence of words into a fixed-dimensional vector~\cite{kim-2014-convolutional}. A linear transformation is applied on a window of $k$ words to produce a set of new features (we set $k=3$ in this work). To avoid that the window size is larger than the sequence length, zero-padding is used on both sides of the input sequence. After processing all possible windows of words, we get a sequence of new feature vectors. A max-over-time pooling operation \cite{collobert2011natural} is applied over new features, which takes the maximum value of each row to form as the final features $\mathbf{c}$. Thus,
\begin{equation}
\text{CNN}(\text{desc}(s))=unit(\mathbf{c})
\end{equation}

\subsubsection{Single Domain and Domain Adaptation}
With one of these three encoders, we can encode the description of a slot as a vector. Thus, the label embedding of tag $t \in \mathcal{T}_d$ is
\begin{equation}
    \text{emb}(t) = [\mathbf{o}(\text{IOB}(t));\text{ENCODER}(\text{desc}(\text{SLOT}(t)))]
\end{equation}
where ENCODER can be mean, BLSTM and CNN, $\text{emb}(t) \in \mathbb{R}^{K}$, $K$ is the embedding length. Note that $\text{emb}(\text{``O''})=\mathbf{0}$.

For any domain $d \in \mathcal{D}_{src} \cup \{td\}$, we get a label embedding matrix, $\mathbf{B}_d \in \mathbb{R}^{K \times |\mathcal{T}_d|}$, and $[\mathbf{B}_d^{\top}]_{t}=\text{emb}(t)$ for any $t \in \mathcal{T}_d$. Both Equation (\ref{eqn:output_layer}) and (\ref{eqn:output_layer_of_domain_adaptation}) can be rewritten as 
\begin{equation}
    \mathbf{z}_i=\mathbf{B}_d^{\top} \mathbf{C} \mathbf{h}_i
\end{equation}
where $\mathbf{z}_i \in \mathbb{R}^{|\mathcal{T}_d|}$, $\mathbf{C} \in \mathbb{R}^{K \times 2n}$ denotes trainable parameters of the output layer, but $\mathbf{B}_d$ is computed by the slot description encoder on domain $d$. Since the encoder is domain-independent, $\mathbf{C}$ and the encoder can also be fully shared across domains, as well as the sentence BLSTM which produces $\mathbf{h}_i$.

This method requires an unstructured short-text to describe each slot, which is easier to be obtained, while the involved label encoders may take more time for computation.

\subsection{Slot Exemplar Encoding}
\label{sec:data_exemplar}

\begin{figure}
\centering
\includegraphics[width=0.48\textwidth]{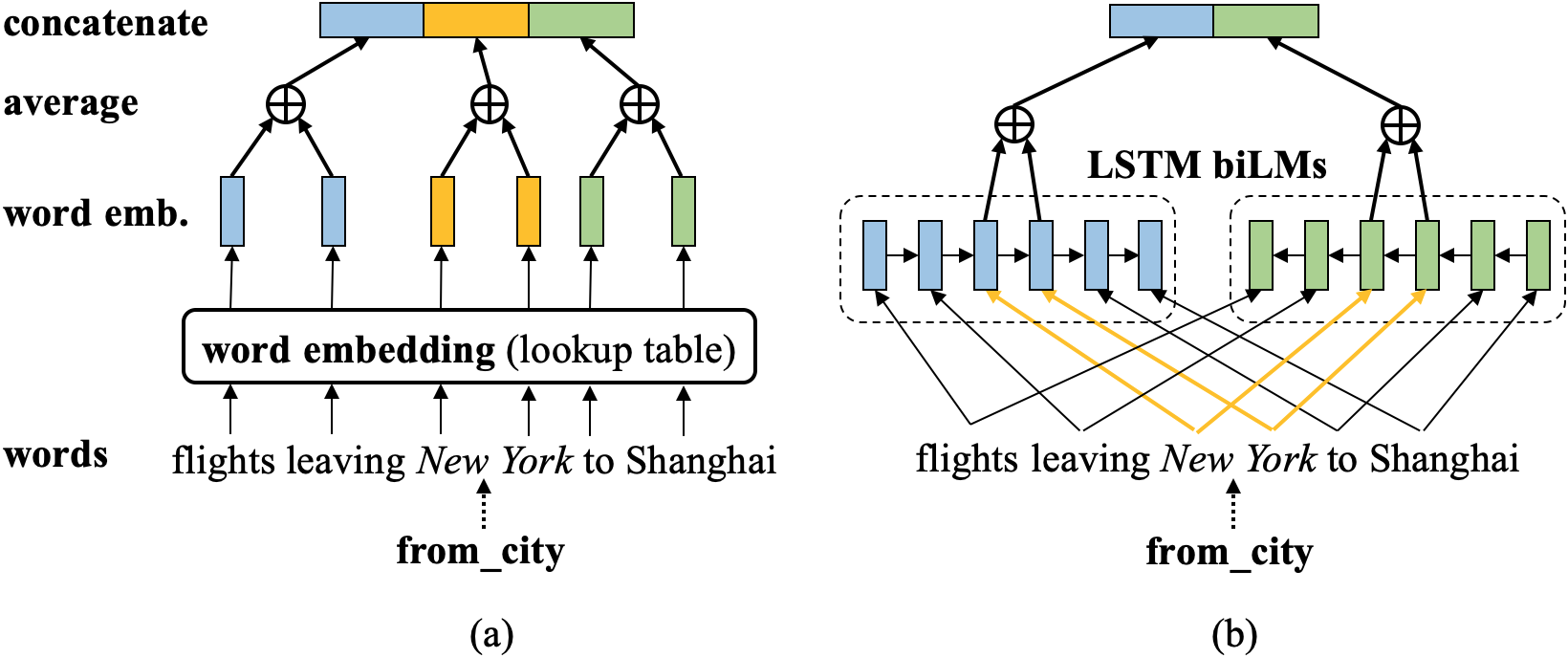}
\caption{Two methods to encode data exemplar of a slot, e.g., {\tt from\_city}. (a) Word embeddings: word embeddings in the left context, value and right context are averaged and concatenated respectively. (b) Language models: a biLMs is used to encode the exemplar into context-aware vectors, and we exploit the vectors at the positions of the slot annotation.}
\label{fig:context_value_embeddings}
\end{figure}

In this method, we propose to extract label embeddings from slot exemplars which contain values of each slot and their neighbor contexts, as shown in Fig. \ref{fig:label_embeddings} (c). An exemplar of slot $s$ can be a sentence $\boldsymbol{e}=e_1\cdots e_{|e|}$ and the ``(\emph{start}, \emph{end})'' positions $(sp,ep)$ for annotation of the slot, i.e. $\boldsymbol{e}_{sp:ep}$ is a value of $s$. For example, one exemplar of slot {\tt from\_city} is ``flights leaving New York to Shanghai'', and ``New York'' at positions $(3, 4)$ is its value. Two methods are introduced to obtain label embedding by encoding slot exemplars, which utilize words embeddings and language models.


\subsubsection{Word Embeddings} Every word in the exemplar sentence is mapped to a vector via $\mathbf{e}_j=\mathbf{W}_{e}\mathbf{o}(e_j)$, where $\mathbf{W}_{e}$ is a word embedding matrix and $\mathbf{o}(e_j)$ is a one-hot vector. To specify the value and context information of the slot, we split the exemplar sentence into three segmentations, left context, value and right context, as illustrated in Fig. \ref{fig:context_value_embeddings} (a). Then, word embeddings in these three parts are respectively averaged and concatenated, composing the slot embedding
\begin{equation}
    \text{emb}(s) = unit([\text{ave}(\mathbf{e}_{<sp});\text{ave}(\mathbf{e}_{sp:ep});\text{ave}(\mathbf{e}_{>ep})])
\end{equation}
where $\text{ave}(\mathbf{e}_{<sp})$ and $\text{ave}(\mathbf{e}_{>ep})$ are zero vectors when $sp=1$ and $ep=|e|$ respectively.

\subsubsection{Language Models} Different from word embeddings, LSTM based bidirectional language models (biLMs)~\cite{peters-etal-2018-deep} directly encode context information into hidden states. The input sentence $\boldsymbol{e}=e_1\cdots e_{|e|}$ is converted into two sequences of hidden vectors ($\overrightarrow{\mathbf{u}}_{1:|e|}, \overleftarrow{\mathbf{u}}_{1:|e|}$) by forward and backward LMs respectively, as shown in Fig. \ref{fig:context_value_embeddings} (b). Since hidden states already contain information of history and current word, we use the hidden vectors at positions ranging from $sp$ to $ep$ to compose the slot embedding:
\begin{equation}
    \text{emb}(s) = unit([\text{ave}(\overrightarrow{\mathbf{u}}_{sp:ep});\text{ave}(\overleftarrow{\mathbf{u}}_{sp:ep})])
\end{equation}


\subsubsection{Single Domain and Domain Adaptation}

Now, we can encode the exemplars of any slot as a vector (if there are multiple exemplars for one slot, an average is taken). Thus, the label embedding of tag $t \in \mathcal{T}_d$ is
\begin{equation}
    \text{emb}(t) = [\mathbf{o}(\text{IOB}(t));\text{emb}(\text{SLOT}(t))]
\end{equation}
where $\text{emb}(t) \in \mathbb{R}^{K}$, $K$ is the embedding length, and $\text{emb}(\text{SLOT}(\text{``O''})) = \mathbf{0}$.


For any domain $d \in \mathcal{D}_{src} \cup \{td\}$, we get the label embedding matrix, $\mathbf{B}_d \in \mathbb{R}^{K \times |\mathcal{T}_d|}$, and $[\mathbf{B}_d^{\top}]_{t}=\text{emb}(t)$ for any $t \in \mathcal{T}_d$. Both Equation (\ref{eqn:output_layer}) and (\ref{eqn:output_layer_of_domain_adaptation}) can be rewritten as 
\begin{equation}
    \mathbf{z}_i=\mathbf{B}_d^{\top} \mathbf{C} \mathbf{h}_i
\end{equation}
where $\mathbf{z}_i \in \mathbb{R}^{|\mathcal{T}_d|}$, $\mathbf{C} \in \mathbb{R}^{K \times 2n}$ denotes trainable parameters of the output layer which can be shared across domains. However, $\mathbf{B}_d$ is computed by using pre-trained word embeddings or biLMs in this paper, which will be fixed.

This method requires minimal prior knowledge which is some kind of labeled data sample, while it may heavily rely on the quality of pre-trained word embeddings or biLMs.

\subsection{Label Embedding-based CRF Layer}
\label{sec:bilinear_transition}

The typical linear-chain CRF provides the transition between a pair of labels as a \emph{scalar} score, as shown in Equation (\ref{eqn:crf_score}). However, prior knowledge about slots is not exploited in the traditional transition matrix of CRF layer. We propose a new transition model of CRF Layer depending on label embeddings for slot filling. It calculates the transition score using a \emph{bilinear} model. The transition score of two tags $t_{i-1}$ and $t_i$ in Equation (\ref{eqn:crf_score}) is computed via:
\begin{equation}
    [\mathbf{A}_d]_{t_{i-1},t_i} = \text{emb}(t_{i-1})^{\top} \mathbf{W}_a \text{emb}(t_{i})
\end{equation}
where $t_{i-1} \in \mathcal{T}_d$, $t_i \in \mathcal{T}_d$, $\mathbf{A}_d \in \mathbb{R}^{|\mathcal{T}_d|\times |\mathcal{T}_d|}$ is a transition matrix for domain $d$. The parameters of $\mathbf{W}_a$ are domain independent and can be fully shared across domains. In the inference stage, $\mathbf{A}_d$ is computed only once per domain.

Compared with the traditional linear-chain CRF whose transition scores are initialized as independent scalars and learned from scratch, we compute transition scores by matching label embeddings of start and end tags. Our model could transfer the knowledge about label transition by utilizing implicit slot relations in label embeddings. Besides the bilinear model, there are other alternatives, like attention mechanism~\cite{bahdanau2014neural,luong2015effective}, multi-head attention~\cite{vaswani2017attention}.

\section{Experiments}
\label{sec:exp}
In this section, we first introduce the datasets and baselines used. Then, Section \ref{sec:exp-setup} gives the details of experimental setup. In section \ref{sec:exp-performance}, we compare the performance of our proposed methods with traditional slot filling methods.

\subsection{Datasets}

We carry out experiments on the following NLU datasets.

\begin{itemize}
    \item \textbf{DSTC 2\&3}~\cite{henderson2013dialog}: It contains two domains, DSTC2 (source domain) and DSTC3 (target domain). DSTC2 comprises of dialogues from the restaurant information domain in Cambridge. DSTC3 introduces a tourist information domain about restaurants, pubs and coffee shops in Cambridge, which is an extension of DSTC2 but has only a seed set for training. We use manual transcripts as input by ignoring speech recognition errors. Several slots of DSTC2 are reused in DSTC3.
    \item \textbf{AICar}: The AICar dataset is a custom dataset collected from a commercial in-car spoken dialogue system which has thousands of users. It consists of three domains: {\tt MAP} navigation, {\tt WEATHER} forecast and {\tt FLIGHT} information. There is almost no slot overlap among these domains, yet the slots are partially associated, e.g., {\tt MAP} has slot {\tt starting\_point}, {\tt WEATHER} has {\tt city\_name}, and {\tt FLIGHT} has {\tt departure\_city}.
    \item \textbf{SNIPS}: It is a public NLU dataset~\cite{2018arXiv180510190C} of crowdsourced user utterances across 7 intents and about 2000 training samples per intent. It contains natural language corpus collected in a crowdsourced fashion to benchmark the performance of voice assistants. 
\end{itemize}

All of the datasets are annotated with slot labels\footnote{Notice that because the original semantic annotations are not aligned with words in DSTC 2\&3, we simply use string matching for alignment, which converts the semantic annotations into the slot filling format.}. The detailed statistics of each dataset are shown in Table \ref{tab:dataset}. The language of DSTC 2\&3 and SNIPS is English, and AICar is in Chinese. To avoid Chinese word segmentation errors~\cite{meng2019word}, the slot filling annotation of AICar is at Chinese-character level.


\begin{table}[t]
  \caption{Dataset statistics. ``AL(desc.)'' denotes the average length of slot descriptions.}
  \label{tab:dataset}%
  \centering{
    \begin{tabular}{|c|c|cc|ccc|}
    \hline
    Dataset  & Domain & Train & Test & \#Slot & \#Atom & AL(desc.) \\
    \hline \hline
    \multirow{2}*{\begin{tabular}[c]{@{}c@{}}DSTC\\ 2\&3\end{tabular}} & DSTC2 & 15611 & 9890 & 17 & 11 & 2.2  \\
     & DSTC3 & 109 & 18715 & 30 & 16 & 2.4  \\
    \hline
    \multirow{3}*{AICar} & Map & 12000 & 20661 & 30 & 16 & 4.3  \\
     & Weather & 10150 & 7236 & 36 & 20 & 3.4  \\
     & Flight & 1500 & 4407 & 22 & 18 & 4.0  \\
    \hline
    SNIPS & - & 13784 & 700 & 39 & - & 1.8 \\
    \hline
    \end{tabular}
  }
\end{table}%

\subsection{Baselines}

We compare with one traditional method, two label embedding methods and two zero-shot learning based approaches: 

\begin{itemize}
    \item One-hot: A 2-layer BLSTM model and one linear output layer (CRF layer is optional) for slot filling, as shown in Section \ref{sec:st_slu}. Each label is discrete and represented as a one-hot vector in this method.
    \item Canonical Correlation Analysis (CCA) label embedding \cite{kim-etal-2015-new}: CCA is exploited to induce slot vector by analysis cooccurrences of slots and words, i.e. words tagged with the corresponding slots.
    \item Value word embedding \cite{ma2016label}: We take the average of pre-trained embeddings of the value words for slot embedding, while ignoring the context words.
    \item Concept Tagger (CT)~\cite{Bapna2017towards} and Zero-shot Adaptive Transfer (ZAT)~\cite{lee2018zero}: These two zero-shot methods also condition slot filling on textual slot descriptions, which have a contextual BLSTM layer and slot dependent BLSTM layer. However, they predict an IOB sequence for each slot, which may lead to overlapped segmentation and cost more time. %
\end{itemize}

\subsection{Experimental Setup}
\label{sec:exp-setup}

\subsubsection{Label embeddings of slot filling}
Three different approaches to extracting label embeddings are investigated in this paper, via atomic concepts, slot descriptions and slot exemplars, which involve different degrees of human knowledge. Table \ref{tab:dataset} also provides the detailed statistics of label representations. For atomic concepts, we act as domain experts to manually split each slot into small units for all five domains, e.g., {\tt deny-starting\_point} is split into ``deny'', ``from\_location'' and ``poi\_name''. As an extension of DSTC2, DSTC3 contains 6 new atoms that constitute about $25\%$ of slot tags. For slot descriptions, we leverage tokenized slot names since each slot name is already a meaningful phrase in the datasets, e.g., {\tt deny-starting\_point} is described as ``deny starting point''. The training data can be directly exploited as data exemplars for the corresponding slots. 

\subsubsection{Network Hyper-parameters}

The experimental setup was kept the same for all three datasets with minor adjustments in hyper-parameters. We adopt a 2-layer BLSTM model to encode sentences with $200$ hidden units. For training, the network parameters are randomly initialized under the uniform distribution $[-0.2, 0.2]$. We use optimizer Adam \cite{kingma2014adam} with learning rate $0.001$ for all experiments. The \emph{dropout} with a probability of $0.5$ is applied to the non-recurrent connections during the training stage. The batch size is $20$ for domain DSTC2, {\tt MAP} and {\tt WEATHER}, $5$ for domain DSTC3 and {\tt FLIGHT}, and $10$ for SNIPS. The maximum norm for gradient clipping is set to 5, and we use weight decay regularization on all weights with lambda=$1e\text{-}6$ to avoid over-fitting.

\subsubsection{Network Input Embeddings}

We exploit pre-trained word embeddings to initialize the input embedding layers. For AICar, we train an LSTM based bidirectional language models (biLMs) at Chinese-character level by using zhwiki\footnote{\url{https://dumps.wikimedia.org/zhwiki/latest}} corpus. The embedding dimension is 200. For DSTC 2\&3 and SNIPS, we directly adopt a pre-trained biLMs of ELMo~\cite{peters-etal-2018-deep} in English\footnote{\url{https://github.com/allenai/allennlp/blob/master/tutorials/how_to/elmo.md}}. The word embedding dimension is 1024. 


\subsubsection{Encoder of slot exemplars}

The pre-trained input embeddings and BiLMs are also exploited to compute label embeddings based on slot exemplars, as shown in Section \ref{sec:data_exemplar}. In this paper, we choose to fix the encoder of slot exemplars to reduce training cost. Therefore, the performance of slot filling based on slot exemplars may heavily rely on qualities of the pre-trained BiLMs.

\subsubsection{Data settings of domain adaptation}
\label{sec:data_setting_of_domain_adaptation}

To evaluate the efficiency of different methods in leveraging data, we set up a domain adaptation problem. In DSTC 2\&3, DSTC2 is source domain, and DSTC3 is target domain. In AICar, we choose each domain as the target one and the rest two domains as source domains. Moreover, $1, 2.5, 5, 10, 20, 40, 60, 80$ percent of the training data in the target domain are randomly selected to simulate low resource settings of the target domain. Each data selection is repeated 10 times by using different random seeds, and the final results of the evaluation are averaged. For SNIPS, we keep one intent as the target domain and the rest intents as the source domain. We pre-train the slot filling model on the source domains, then fine-tune and evaluate it on the target domain.

We randomly selected $80\%$ of the training data for model training and the remaining $20\%$ for validation. We keep the learning rate for 50 epochs and save the parameters that give the best performance on the validation set. Finally, we report the $\text{F}_1$-score of the semantic slots on the test set with parameters that have achieved the best $\text{F}_1$-score on the validation set. The $\text{F}_1$-score is calculated using CoNLL evaluation script\footnote{\url{https://www.clips.uantwerpen.be/conll2000/chunking/output.html}}.

\subsubsection{Significance Test} We use McNemar's test to establish the statistical significance of a method over another ($\text{p}<0.05$).

\subsection{Main Results}
\label{sec:exp-performance}

\begin{table}[]
  \caption{Slot F$_1$ scores on DSTC3 test set with or without the training data of DSTC2. The bold number indicates a statistically significant gain over the best baseline.}
  \label{tab:dstc23_adaptation}%
  \centering{
    \begin{tabular}{|l||c|c|}
    \hline
    \textbf{Method} & \begin{tabular}[c]{@{}c@{}}DSTC3\_train\\ (single domain)\end{tabular} & \begin{tabular}[c]{@{}c@{}}+DSTC2\_train\\ (domain adapt.)\end{tabular} \\
    \hline \hline
    One-hot & 75.52 & 83.14 \\
    \ \ \ \ (-) w/o output layer sharing & 75.52 & 81.22 \\
    \hline
    CCA label embedding & 76.25 & 84.10 \\
    value word embedding & 75.32 & 83.97 \\
    CT & 76.20 & 83.68 \\
    ZAT & 75.76 & 84.29 \\
    \hline
    atomic concept & 76.41 & \textbf{86.03} \\
    slot description (BLSTM) & \textbf{77.41} & \textbf{87.16} \\
    slot description (CNN) & 76.23 & \textbf{85.89} \\
    slot description (emb mean) & 76.56 & 84.05 \\
    slot exemplar (word embedding) & \textbf{77.19} & \textbf{86.10} \\
    slot exemplar (BiLMs) & \textbf{78.58} & \textbf{87.55} \\
    \hline
    \end{tabular}%
  }
\end{table}%

\begin{table*}[t]
  \caption{Slot F$_1$ scores on AICar dataset with three domains. We train and evaluate on each target domain and pre-train the slot filling model on the rest two source domains. Columns represent different proportions of the training data in the target domain, defined in Section \ref{sec:data_setting_of_domain_adaptation}. The results in bold black are the best F$_1$ scores. Time per batch refers to training and testing times on GeForce GTX 1080 Ti GPU.}
  \label{tab:navi_weather_flight_adaptation_all}%
  \centering{
    \begin{tabular}{|c|l||l|cccccccccc|}
    \hline
    \begin{tabular}[c]{@{}c@{}}\textbf{Target}\\ \textbf{Domain}\end{tabular} & \textbf{Method} & \begin{tabular}[c]{@{}l@{}}\textbf{Time per}\\ \textbf{batch}\end{tabular} & 0\% & 1\% & 2.5\% & 5\% & 10\% & 20\% & 40\% & 60\% & 80\% & 100\%  \\
    \hline \hline
    \multirow{9}*{Map} & One-hot & 9.2ms & 0 & 40.81 & 60.24 &	72.67 &	82.63 &	87.94 &	90.99 &	92.92 &	93.61 &	94.79   \\
     & CT & 94.2ms & 0.22 & 40.93 & 62.23 & 75.20 & 82.79 & 87.58 & 91.28 & 92.88 & 94.02 & 94.80 \\
     & ZAT & 118ms & 0.05 & 45.07 & 63.74 & 73.79 & 82.10 & 87.39 & 91.33 & 92.40 & 93.65 & 94.62 \\
    \cline{2-13}
     & atomic concept & 9.5ms & \textbf{1.55} & 46.34 & 66.57 & 77.20 & 84.74 & 89.27 & 92.04 & 93.76 & \textbf{94.81} & \textbf{95.78} \\
     & slot description (BLSTM) & 11.9ms & 0.89 & \textbf{46.55} & 67.26 & 77.11 & 84.83 & 88.68 & 92.02 & 93.81 & 94.56 & 95.65 \\
     & slot description (CNN) & 10.4ms & 0.78 & 43.44 & 67.41 & 77.88 & 84.85 & 88.79 & 91.86 & 93.77 & 94.41 & 95.31 \\
     & slot description (emb mean) & 9.9ms & 0.32 & 45.20 & \textbf{70.12} & \textbf{79.95} & \textbf{85.92} & \textbf{89.37} & \textbf{92.48} & \textbf{93.90} & 94.67 & 95.47 \\
     & slot exemplar (word embedding) & 9.2ms & 0 & 42.54 & 64.34 & 76.29 & 84.16 & 88.45 & 91.61 & 93.40 & 94.10 & 94.71 \\
     & slot exemplar (BiLMs) & 9.2ms & 0 & 45.49 & 66.17 & 76.74 & 84.45 & 88.80 & 91.91 & 93.61 & 94.42 & 95.72 \\
        \hline \hline
    \multirow{9}*{Weather} & One-hot & 7.5ms & 0 & 32.91 & 39.35 &    49.23    & 62.40 & 72.96 & 80.57 & 84.13 & 86.50 & 88.60   \\
     & CT & 140ms & 4.97 & 39.47 & 45.46 &    52.26    & 62.67 & 73.39 & 81.42 & 84.85 & 86.96 & 87.85 \\
     & ZAT & 187ms & 15.73 & 39.78 & 47.03 & 54.10 & 65.28 & 73.60 & 80.60 & 84.19 & 86.75 & 88.61 \\
    \cline{2-13}
     & atomic concept & 8.0ms & \textbf{24.69} & 41.73 & \textbf{48.38} &    56.51    & 67.39 & 75.67 & 82.48 & 85.08 & 87.35 & 88.79 \\
     & slot description (BLSTM) & 10.8ms & 6.81 & \textbf{42.14} & 47.41 &    \textbf{56.67}    & 66.84 & 75.72 & 82.70 & \textbf{85.89} & \textbf{88.23} & 89.43 \\
     & slot description (CNN) & 9.3ms & 16.74 & 40.08 & 46.47 &    54.73    & \textbf{67.41} & 75.85 & \textbf{82.74} & 85.37 & 87.86 & \textbf{89.44} \\
     & slot description (emb mean) & 8.4ms & 15.75 & 41.72 & 48.32 &    55.71    & 66.54 & \textbf{76.06} & 82.47 & 85.56 & 87.90 & 89.42 \\
     & slot exemplar (word embedding) & 7.7ms & 0 & 41.41 & 46.59 &    55.09    & 65.73 & 75.28 & 82.44 & 85.31 & 87.56 & 88.72 \\
     & slot exemplar (BiLMs) & 7.8ms & 0 & 39.35 & 44.28 &    52.41    & 64.50 & 74.61 & 81.62 & 84.75 & 87.56 & 88.00 \\
         \hline \hline
    \multirow{9}*{Flight} & One-hot & 6.2ms & 0 & 55.69 & 65.84 & 73.22 & 80.68 & 87.58 & 92.26 & 94.14 & 94.92 & 95.26   \\
     & CT & 69.5ms & 2.07 & 59.31 & 71.01 & 75.36 & 81.40 & 87.74 & 92.32 & 93.84 & 94.97 & 95.74 \\
     & ZAT & 99.2ms & 1.89 & 65.91 & 70.44 & 74.80 & 82.03 & 87.81 & 92.37 & 93.93 & 94.77 & 95.34 \\
    \cline{2-13}
     & atomic concept & 6.7ms & \textbf{52.37} & 62.72 & 72.74 & 78.58 & 83.42 & 88.76 & 92.79 & 93.99 & 95.02 & 95.27 \\
     & slot description (BLSTM) & 9.3ms & 2.95 & \textbf{70.49} & 75.93 & 79.93 & 85.49 & 89.96 & 92.94 & \textbf{94.99} & 95.33 & 95.47 \\
     & slot description (CNN) & 8.4ms & 2.41 & 64.78 & 73.15 & 78.45 & 84.92 & 89.58 & 93.17 & 94.69 & \textbf{95.41} & 95.83 \\
     & slot description (emb mean) & 7.0ms & 8.45 & 69.97 & \textbf{76.02} & \textbf{80.77} & \textbf{85.79} & \textbf{90.33} & \textbf{93.33} & 94.60 & 95.29 & 95.76 \\
     & slot exemplar (word embedding) & 6.4ms & 0 & 68.49 & 73.73 & 79.13 & 83.82 & 88.93 & 92.56 & 94.18 & 94.95 & 95.30 \\
     & slot exemplar (BiLMs) & 6.3ms & 0 & 66.58 & 72.96 & 77.85 & 82.93 & 88.66 & 92.92 & 94.39 & 95.17 & \textbf{95.98} \\
    \hline
    \end{tabular}%
  }
\end{table*}%

We first compare different label embeddings by exploiting BLSTM-\emph{softmax} models for slot filling, except for CT and ZAT. BLSTM-CRF models with the \emph{scalar} and \emph{biliner} transition modules are then investigated in Section \ref{sec:exp_with_crf_layer}.

The results on DSTC 2\&3 dataset are shown in Table \ref{tab:dstc23_adaptation}. By comparing the column ``DSTC3\_train'' with ``+DSTC2\_train'', we can see performances on the target domain are improved dramatically due to that the parameters are pre-trained on the source domain. Row 1-2 show that sharing the output weights of reused labels (see Equation (\ref{eqn:output_layer_sharing})) can help transfer knowledge from DSTC2 to DSTC3. By comparing the baselines with our proposed methods, most of our methods can outperform all baselines on both single domain and domain adaptation tasks. Notably, by using label embeddings extracted from slot exemplars with the pre-trained biLMs of ELMo, the slot filling model can achieve the best results ($78.58\%$ and $87.55\%$). 

Table \ref{tab:navi_weather_flight_adaptation_all} shows the performances of domain adaptation on AICar dataset with three domains. We take each domain as the target domain one by one and leave the rest two as the source domains. We evaluate the baselines and our methods with parts of the training data on the target domain, which range from 1 to 100 percent. We also test the pre-trained models even without the target training data (zero-shot). From Table \ref{tab:navi_weather_flight_adaptation_all}, we can find that:

\begin{itemize}[\IEEEsetlabelwidth{Z}]
    \item Two zero-shot baselines (CT and ZAT) perform better than the One-hot, which benefit from slot descriptions. However, these slot-independent models increase the computation cost sharply since they need to predict an IOB sequence for each slot independently. This could also lead to overlapped slot segmentations and lose the time series dependence of output slots. 
    \item Even without the target training data ($0\%$), our methods outperform all the baselines. Especially, atomic concepts can achieve the best performances. Because atomic concepts give an explicit interpretation of slots, which can recombine associated atoms into a new slot. While slot descriptions as word sequences are more complex and may contain synonyms implicitly. Without the target training data, there is no exemplar available for each slot.
    \item With the increasing training data of the target domain, slot $\text{F}_1$ scores of all methods are improved concurrently.
    \item At all data partitions, our proposed methods can achieve the best results and achieve statistically significant gain over the best baseline in most cases. The best performances always lie on atomic concepts and slot descriptions, since they involve more human knowledge than slot exemplars which are just the available training samples. Even so, the slot exemplar-based method can also outperform the baselines in most cases.
\end{itemize}

\subsubsection{Input embeddings with pre-trained language model} Besides using pre-trained word embeddings, some advanced pre-trained language models (e.g. ELMo~\cite{peters-etal-2018-deep}, BERT~\cite{devlin2018bert}) can also be used to get input embeddings. However, it is orthogonal to the investigation of label embeddings. Table \ref{tab:dstc23_adaptation_elmo_bert} shows results on DSTC3 test set using limited seed instances, indicating that our method can get improvements over different types of input embeddings simultaneously. For BERT, \emph{bert-base-cased} model\footnote{\url{https://github.com/google-research/bert#pre-trained-models}} is utilized, while it performs worse than ELMo. The reason may be that ELMo could get better generalization for the dataset by using char-based word embeddings.

\begin{table}[]
  \caption{Slot F$_1$ scores on DSTC3 test set without the training data of DSTC2, but with different pre-trained language models.}
  \label{tab:dstc23_adaptation_elmo_bert}%
  \centering{
    \begin{tabular}{|c||c|c|c|}
    \hline
    \textbf{Method} & pre. word emb. & ELMo & BERT \\
    \hline \hline
    One-hot & 75.52 & 78.13 & 76.64 \\
    \hline
    slot description (BLSTM) & 77.41 & 80.03 & 78.13 \\
    \hline
    \end{tabular}%
  }
\end{table}

\subsubsection{Learning curve} 
\begin{figure}[]
\centering
\includegraphics[width=0.4\textwidth]{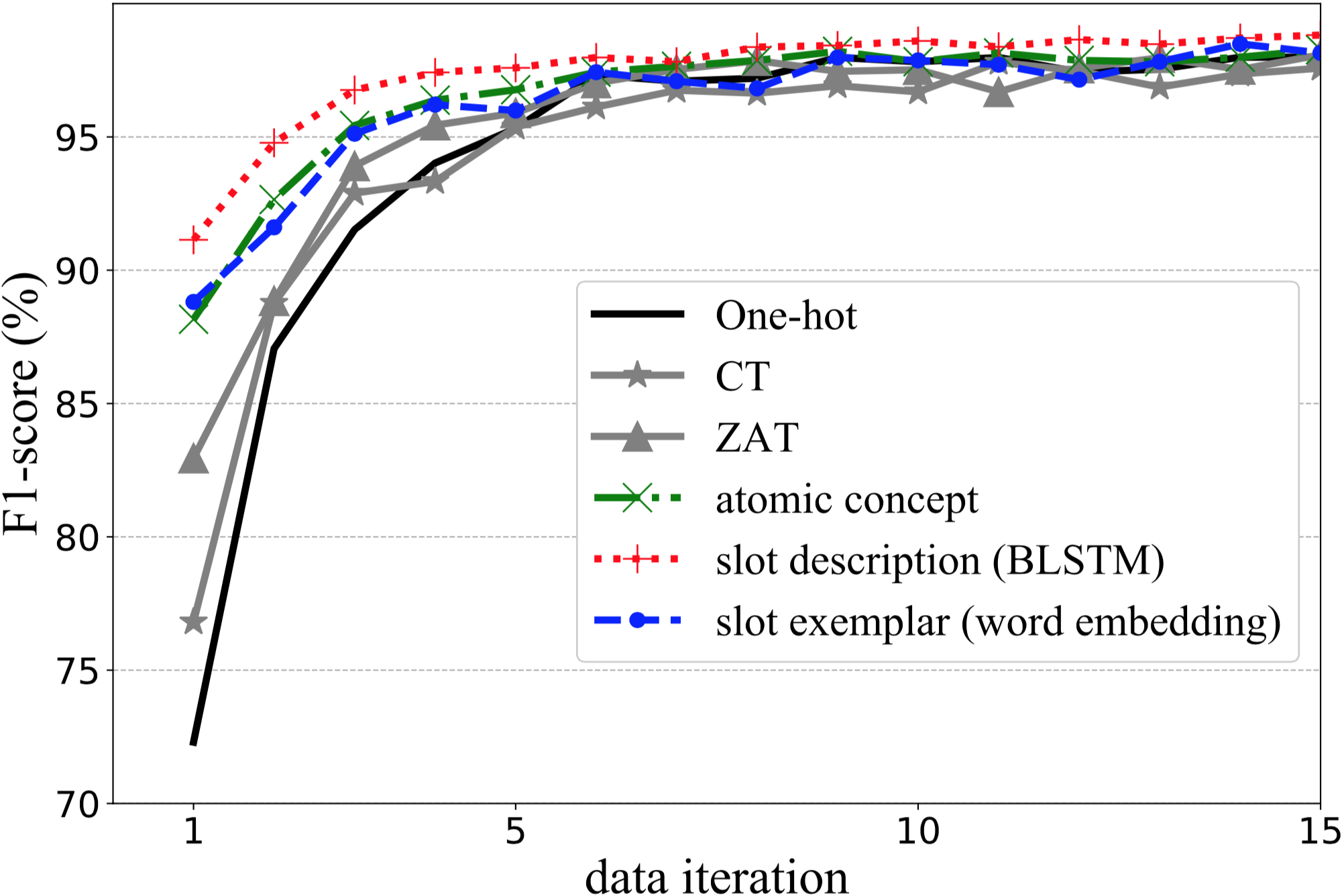}
\caption{Learning curves of the baselines and our methods when {\tt FLIGHT} is the target domain and 100\% training data is used for fine-tuning. It shows performances on the validation set at different data iterations.}
\label{fig:flight_learning_curve}
\end{figure}

Fig. \ref{fig:flight_learning_curve} shows learning curves of different methods when the target domain is {\tt FLIGHT} and 100\% training data is used for fine-tuning. It shows slot $\text{F}_1$ scores on the validation set at increasing numbers of data iteration. We can see that our methods can converge much faster than the baselines without label embeddings. Especially, the ``slot description (BLSTM)'' method converges fastest.

\subsubsection{Number of slot exemplars} 
\begin{figure}[]
\centering
\includegraphics[width=0.4\textwidth]{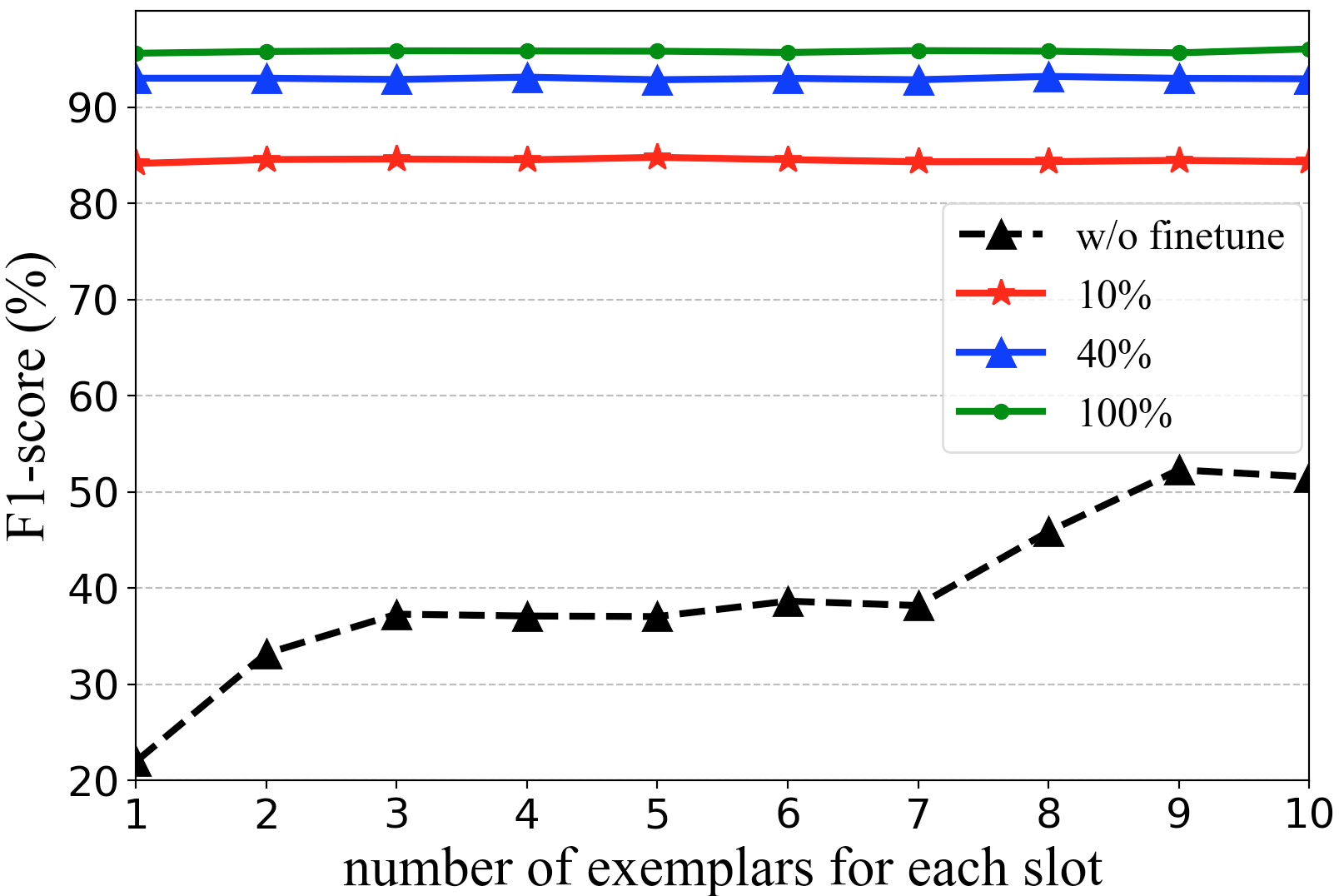}
\caption{Performances of the ``slot exemplar (BiLMs)'' method when {\tt FLIGHT} is the target domain and $\{\text{none}, 10\%, 40\%, 100\%\}$ training data is used for fine-tuning. The number of exemplars for each slot ranges from 1 to 10.}
\label{fig:flight_exemplar_number}
\end{figure}

Although we exploit all data samples of the training set as data exemplars for the corresponding slots, it is meaningful to investigate that how many exemplars for each slot are enough. Fig. \ref{fig:flight_exemplar_number} shows the number of slot exemplars used versus performance on {\tt FLIGHT} domain. If there is no training data for fine-tuning, more exemplars usually give better performance. If there is an amount of labelled data for training, the performance is dominated by the size of training data in the target domain.

\subsubsection{CRF layer}\label{sec:exp_with_crf_layer} We also apply a CRF layer to model the time series dependence of output labels, which CT and ZAT cannot model. We want to know whether label embeddings can improve different slot filling models. Table \ref{tab:navi_weather_to_flight_with_crf} shows the results of using different label embeddings with or without the CRF layer, in the case that {\tt FLIGHT} is the target domain and $10\%$ training data is used for fine-tuning. From the table, we can see that CRF layers give improvements for the baseline and our methods. Compared to the traditional \emph{scalar} transition model (as shown in Equation (\ref{eqn:crf_score})), our proposed \emph{bilinear} transition model (as shown in Section \ref{sec:bilinear_transition}) can improve significantly. The \emph{bilinear} model calculates the transition score of two labels depending on their label embeddings, which shares transition patterns of related labels. It helps mitigate the data sparsity problem. 

\subsubsection{Discussions about varied slot descriptions} Different from atomic concepts or slot exemplars which can be in a united form, slot descriptions in natural language would be varied due to different domain experts or developers. Five participants were asked to describe each slot in the {\tt FLIGHT} domain. The average length of slot descriptions from different participants varies from $6.2$ to $7.7$. If we use these slot descriptions in the ``slot description (BLSTM)'' method when {\tt FLIGHT} is target domain and $10\%$ training data can be used for fine-tuning, slot $\text{F}_1$ scores range from $84.66\%$ to $86.43\%$. As shown in Table \ref{tab:varied_slot_descriptions}, our method seems robust to varied slot descriptions.

\subsubsection{Results on SNIPS dataset} We also report results on the SNIPS public dataset where one intent are kept as the target domain and the rest intents serve as the source domain. Only 50 train instances of the target domain can be utilized for fine-tuning, which are chosen in random. All 7 intents play a role of the target domain one by one, and the slot $\text{F}_1$ scores are averaged over seven target domains. As shown in Table \ref{tab:snips_results}, our methods can also outperform the baselines. Especially, the ``slot exemplar (BiLMs)'' method achieves the best slot $\text{F}_1$ score.

\begin{table}[t]
  \centering{
\begin{threeparttable}[b]
  \caption{Slot $\text{F}_1$ scores of different models with and without CRF layers on AICar dataset, when {\tt FLIGHT} is the target domain and the rest two are the source domains. $10\%$ training data of {\tt FLIGHT} is used for fine-tuning, and results are averaged over 10 random data splits.}
  \label{tab:navi_weather_to_flight_with_crf}%
    \begin{tabular}{|l||c|c|c|}
    \hline
    \multirow{2}*{\textbf{Method}} & \multirow{2}*{w/o CRF} & \multicolumn{2}{c|}{with CRF} \\
    \cline{3-4}
    & & \emph{scalar} & \emph{bilinear} \\
    \hline \hline
    One-hot & 80.68 & 82.74 & 85.49\tnote{a} \\
    \hline
    atomic concept & 83.42 & 84.26 & 86.95 \\
    slot description (BLSTM) & 85.49 & \textbf{86.83} & \textbf{88.46} \\
    slot description (CNN) & 84.92 & 85.56 & 87.24 \\
    slot description (emb mean) & \textbf{85.79} & 86.76 & 87.49 \\
    slot exemplar (word embedding) & 83.82 & 85.61 & 87.10 \\
    slot exemplar (BiLMs) & 82.93 & 84.45 & 86.83 \\
    \hline
    \end{tabular}%
  \begin{tablenotes}
    \item [a] weights of the linear output layer act as label embeddings in the \emph{bilinear} transition model.
  \end{tablenotes}
\end{threeparttable}
  }
\end{table}%

\begin{table}[t]
  \caption{Performances of the ``slot description (BLSTM)'' method with varied slot descriptions on {\tt FLIGHT}. {\tt FLIGHT} is the target domain, and its $10\%$ training data is used for fine-tuning. ``AL(desc.)'' denotes the average length of slot descriptions.}
  \label{tab:varied_slot_descriptions}%
  \centering{
    \begin{tabular}{|c||c|c|c|c|c|c|}
    \hline
    \textbf{AL(desc.)} & 4.0 (ori.) & 7.7 & 7.4 & 7.0 & 6.4 & 6.2 \\
    \hline \hline
    \textbf{Slot $\text{F}_1$} & 85.49 & 85.84 & 84.66 & 85.80 & 86.43 & 85.08    \\
    \hline
    \end{tabular}%
  }
\end{table}%

\begin{table}[t]
  \caption{Slot F$_1$ scores on SNIPS test set in the setting of domain adaptation with only 50 instances of each target intent for fine-tuning.}
  \label{tab:snips_results}%
  \centering{
    \begin{tabular}{|l||c|}
    \hline
    \textbf{Method} & Average Slot F$_1$ \\
    \hline \hline
    One-hot & 85.63   \\
    CT & 83.79  \\
    ZAT & 86.10  \\
    \hline
    slot description (BLSTM) & 86.94 \\
    slot description (CNN) & 86.99  \\
    slot description (emb mean) & 86.57 \\
    slot exemplar (word embedding) & 85.67 \\
    slot exemplar (BiLMs) & \textbf{87.33} \\
    \hline
    \end{tabular}%
  }
\end{table}%

\section{Conclusion}
\label{sec:conclusion}

This paper has proposed prior knowledge-driven label embedding for the NLU slot filling task to address the data sparsity problem (especially label space data sparsity). It requires human knowledge about slot definitions to consider slot relations into the label embeddings. We investigate three different kinds of prior knowledge about slots and the corresponding encoding methods: atomic concepts, slot descriptions and slot exemplars. These label embeddings are fused with different traditional slot filling models (e.g., BLSTM-\emph{softmax} and BLSTM-CRF). Notably, we propose a novel label transition model based on label embeddings for CRF. The proposed approaches are evaluated on three datasets (DSTC 2\&3, AICar and SNIPS) in the settings of single domain and especially domain adaptation. From the experimental results, we find that these label embeddings involving prior knowledge can significantly improve the performances and efficiencies over traditional slot filling models.

The proposed label embedding based slot filling framework shows promising perspectives of future improvements:
\begin{itemize}
    \item The slot exemplar encoding relies on a pre-trained and high qualified BiLM which is fixed. In future work, we will develop a trainable encoder for slot exemplars.
    \item There are a few other kinds of prior knowledge, like the hierarchical dependency of \emph{domain-intent-slot} \cite{lee2018coupled}. Graph neural networks (GNN) have been applied to encode structured data \cite{chen-etal-2018-structured,chen2019agentgraph}. Applying GNN to encode structured knowledge is one of the future work.
    \item Nowadays, developing NLU systems for certain narrow domains independently is the mainstream approach. It costs much to collect data, while it is convenient for domain experts or developers to provide a few domain definitions (prior knowledge). Combination of deep learning-based slot filling and prior knowledge of narrow domains will be a promising research direction.
\end{itemize}

\ifCLASSOPTIONcaptionsoff
  \newpage
\fi



%

\bibliographystyle{IEEEtran}
\bibliography{ref}

%

\begin{IEEEbiography}[{\includegraphics[width=1in,height=1.25in,clip,keepaspectratio]{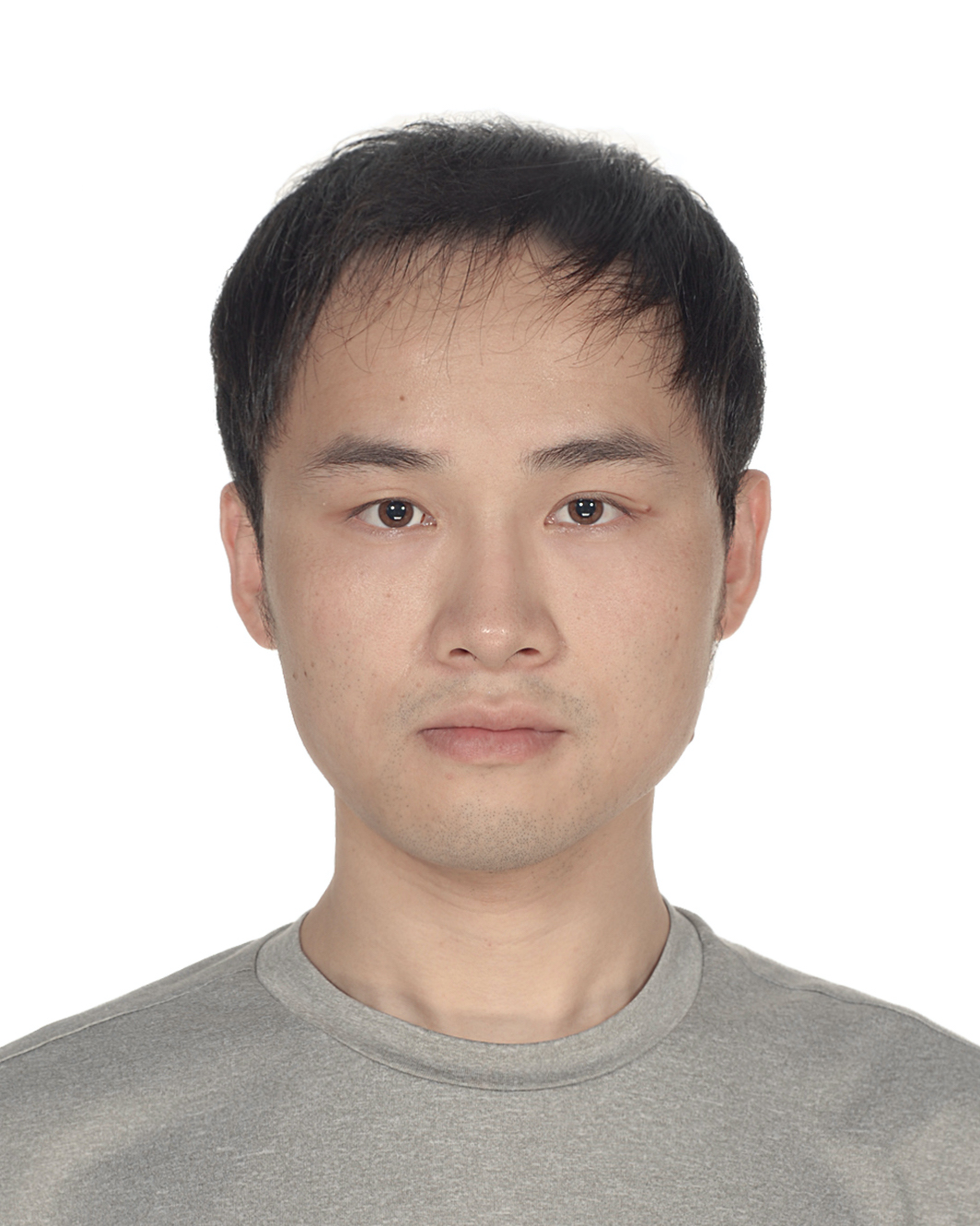}}]{Su Zhu}
received the B.Eng. degree in computer science from Xi’an Jiao Tong University, China in 2013, and the  M.Sc. degree from the Department of Computer Science, Shanghai Jiao Tong University, Shanghai, China, in 2016. He is currently working toward the Ph.D. degree at the SpeechLab, Department of Computer Science and Engineering, Shanghai Jiao Tong University, Shanghai, China. His research interests include spoken/natural language understanding, dialogue systems, and structured deep learning.
\end{IEEEbiography}

\begin{IEEEbiography}[{\includegraphics[width=1in,height=1.25in,clip,keepaspectratio]{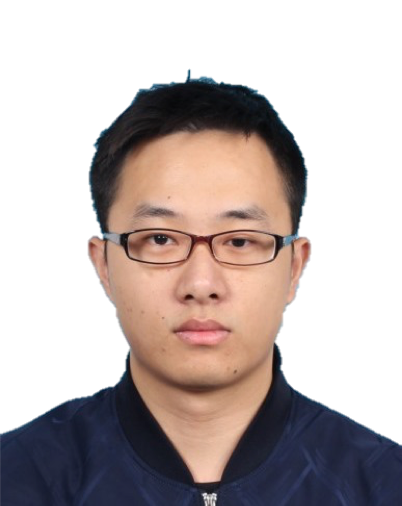}}]{Zijian Zhao}
received the B.Eng. degree in computer science from Shanghai Jiao Tong University, Shanghai, China, in 2017. He is currently working toward the M.S. degree with the SpeechLab, Department of Computer Science and Engineering, Shanghai Jiao Tong University, Shanghai, China. His research interests include dialogue system, spoken language understanding, and machine learning.
\end{IEEEbiography}

\begin{IEEEbiography}[{\includegraphics[width=1in,height=1.25in,clip,keepaspectratio]{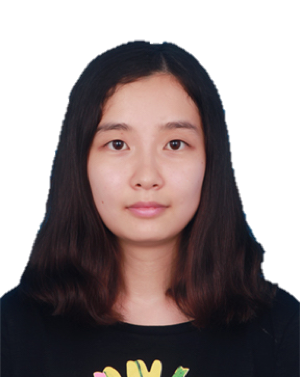}}]{Rao Ma}
received the B.Eng. degree in computer science and technology from Nanjing University, China in 2018. She is currently working toward the M.S. degree with the SpeechLab, Department of Computer Science and Engineering, Shanghai Jiao Tong University, Shanghai, China. Her research interests include natural language processing, language model and large vocabulary continuous speech recognition.
\end{IEEEbiography}


\begin{IEEEbiography}[{\includegraphics[width=1in,height=1.25in,clip,keepaspectratio]{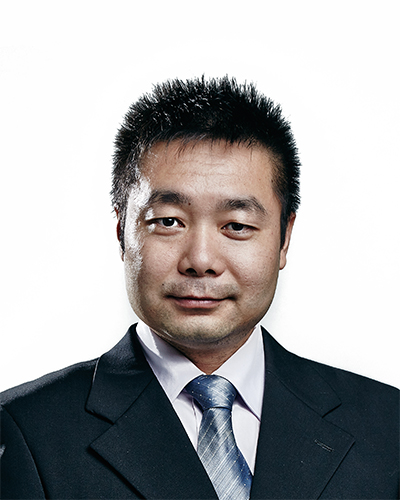}}]{Kai Yu}
is a professor at Computer Science and Engineering Department, Shanghai Jiao Tong University, China. He received his B.Eng. and M.Sc. from Tsinghua University, China in 1999 and 2002, respectively. He then joined the Machine Intelligence Lab at the Engineering Department at Cambridge University, U.K., where he obtained his Ph.D. degree in 2006. His main research interests lie in the area of speech-based human machine interaction including speech recognition, synthesis, language understanding and dialogue management. He is a member of the IEEE Speech and Language Processing Technical Committee.
\end{IEEEbiography}




\end{document}